\documentclass[10pt,twocolumn,letterpaper]{article}

\usepackage{cvpr}              
\usepackage[accsupp]{axessibility}  
\usepackage{graphicx}
\usepackage{amsmath}
\usepackage{amssymb}
\usepackage{booktabs}

\usepackage{lipsum}
\usepackage{newfloat}
\usepackage{listings}
\usepackage{helvet}
\usepackage{courier}
\usepackage{multirow}
\usepackage{color}
\usepackage{subcaption}
\usepackage{verbatim}
\usepackage{url}
\usepackage{textcomp}
\usepackage{gensymb}
\usepackage{graphics} 
\usepackage{epsfig} 
\usepackage{mathptmx} 
\usepackage{times} 
\usepackage{amsmath} 
\usepackage{amssymb}  
\usepackage[table]{xcolor}
\definecolor{gain}{HTML}{34a853}  %
\newcommand{\red}[1]{{\color{red} #1}}

\newcommand{\black}[1]{{\color{black} #1}}

\newcommand{\res}[2]{{$\bm{#1} $ } {({\footnotesize \black{#2}})}}
\newcommand{\ress}[2]{{$\bm{#1} $} {({\footnotesize \black{#2}})}}
\newcommand{\Name}{{Bi-LRFusion}\xspace}
\usepackage{subcaption}
\definecolor{Gray1}{rgb}{0.92,0.92,0.92}
\usepackage{bm}
\usepackage[pagebackref,breaklinks,colorlinks]{hyperref}

\usepackage[capitalize]{cleveref}
\crefname{section}{Sec.}{Secs.}
\Crefname{section}{Section}{Sections}
\Crefname{table}{Table}{Tables}
\crefname{table}{Tab.}{Tabs.}

\def\cvprPaperID{2435} 
\def\confName{CVPR}
\def\confYear{2023}

\begin{document}

\title{\Name: Bi-Directional LiDAR-Radar Fusion for 3D  Dynamic \\Object Detection}

\author{Yingjie Wang$^{1,4}$\hspace{0.15cm} Jiajun Deng$^{2\ast}$\hspace{0.15cm} Yao Li$^{1}$\hspace{0.15cm} Jinshui Hu$^{3}$\hspace{0.15cm} Cong Liu$^{3}$\hspace{0.15cm} Yu Zhang$^{1}$\hspace{0.15cm}  Jianmin Ji$^{1}$ \\Wanli Ouyang$^{4}$\hspace{0.15cm} Yanyong Zhang$^{1\ast}$\hspace{0.15cm} \\
\normalsize $^1$University of Science and Technology of China \hspace{0.5cm} $^2$University of Sydney \hspace{0.5cm} $^3$iFLYTEK \hspace{0.5cm} $^4$Shanghai AI Laboratory\\
}

\maketitle
\newcommand\blfootnote[1]{%
\begingroup
\renewcommand\thefootnote{}\footnote{#1}%
\addtocounter{footnote}{-1}%
\endgroup
}
\begin{abstract}
LiDAR and Radar are two complementary sensing approaches in that LiDAR specializes in capturing an object's 3D shape while Radar provides longer detection ranges as well as velocity hints. 
Though seemingly natural, how to efficiently combine them for improved feature representation is still unclear. The main challenge arises from that Radar data are extremely sparse and lack height information. Therefore, directly integrating Radar features into LiDAR-centric detection networks is not optimal.
In this work,
 we introduce a bi-directional LiDAR-Radar fusion framework, termed \Name, to tackle the challenges and  improve 3D detection for dynamic objects. Technically, \Name involves two steps: first, it enriches Radar's local features by learning important details from the LiDAR branch to alleviate the problems caused by the absence of height information and extreme sparsity; second, it combines LiDAR features with the enhanced Radar features in a unified bird’s-eye-view representation. 
We conduct extensive experiments on nuScenes and ORR datasets, and show that our Bi-LRFusion achieves state-of-the-art performance for detecting dynamic objects. Notably, Radar data in these two datasets have different formats, which demonstrates the generalizability of our method. Codes are available at \url{https://github.com/JessieW0806/Bi-LRFusion}.

\end{abstract}
\blfootnote{*Corresponding Author: Jiajun Deng and Yanyong Zhang.}
\section{Introduction}
\label{intro}
\begin{figure}[ht!]
\centering
\includegraphics[width=0.47\textwidth]{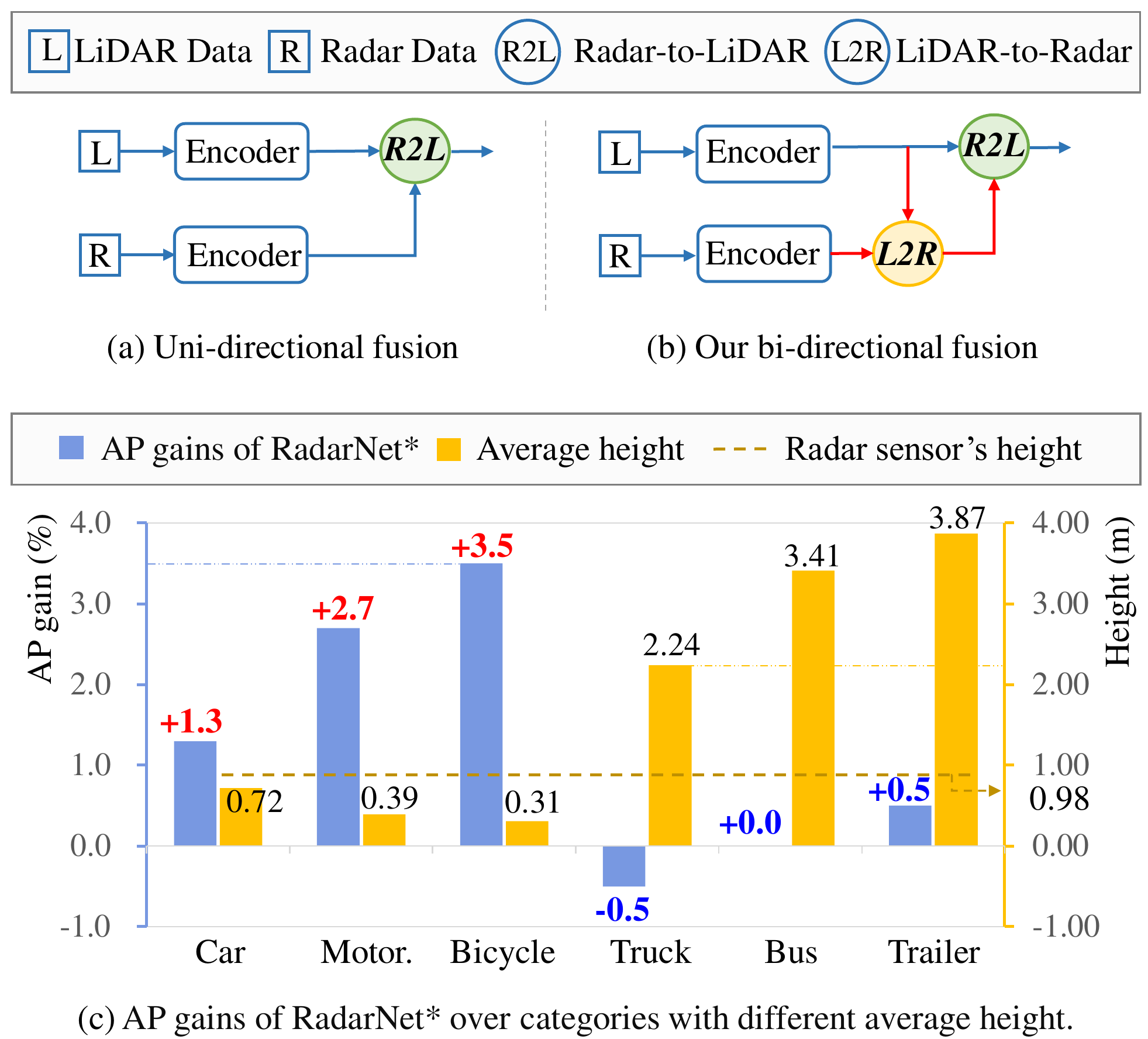}

\caption{An illustration of (a) uni-directional LiDAR-Radar fusion mechanism, (b) our proposed bi-directional LiDAR-Radar fusion mechanism, and (c) the average precision gain (\%) of uni-directional fusion method RadarNet$^*$ against the LiDAR-centric baseline CenterPoint~\cite{yin2021center} over categories with different average height (m). We use * to indicate it is re-produced by us on the CenterPoint.
The improvement by involving Radar data is not consistent
for objects with different height, \emph{i.e.,} taller objects like truck, bus and trailer do not enjoy as much performance gain. 
Note that all height values are transformed to the LiDAR coordinate system.
}

\label{fig:problem}
\end{figure}

LiDAR has been considered as the primary sensor in the perception subsystem of most autonomous vehicles (AVs) due to its capability of providing accurate position measurements~\cite{wang2021multi,deng2021multi,li2022homogeneous}. However, in addition to object positions, AVs are also in an urgent need for estimating the motion state information (\emph{e.g.}, velocity), especially for dynamic objects. Such information cannot be measured by LiDAR sensors since they are insensitive to motion. As a result, millimeter-wave Radar (referred to as Radar in this paper) sensors are engaged  because they are able to infer the object's relative radial velocity~\cite{nabati2021centerfusion} based upon the Doppler effect~\cite{sless2019road}. Besides, on-vehicle Radar usually offers longer detection range than LiDAR~\cite{yang2020radarnet}, which is particularly useful on highways and expressways.
In the exploration of combining LiDAR and Radar data for ameliorating 3D dynamic object detection, the existing approaches~\cite{nobis2021radar,yang2020radarnet,qian2021robust} follow the common mechanism of \textit{uni-directional} fusion, as shown in Figure~\ref{intro} (a). Specifically, these approaches directly utilize the Radar data/feature to enhance the LiDAR-centric detection network without first improving the quality of the feature representation of the former.

However, independently extracted Radar features are not enough for refining LiDAR features, since Radar data are extremely sparse and lack the height information\footnote{This shortcoming is due to today's Radar technology, which may likely change as the technology advances very rapidly, \emph{e.g.,} new-generation 4D Radar sensors~\cite{brisken2018recent}.}.
Specifically, taking the data from the nuScenes dataset~\cite{caesar2020nuscenes} as an example, the 32-beam LiDAR sensor produces approximately 30,000 points, while the Radar sensor only captures about 200 points for the same scene. The resulting Radar bird's eye view (BEV) feature hardly attains valid local information after being processed by local operators (\emph{e.g.}, the neighbors are most likely empty when a non-empty Radar BEV pixel is convolved by convolutional kernels). Besides, on-vehicle Radar antennas are commonly arranged horizontally, hence missing the height information in the vertical direction. 
In previous works, the height values of the Radar points are simply set as the ego Radar sensor's height. Therefore, when features from Radar are used for enhancing the feature of LiDAR, the problematic height information of Radar leads to unstable improvements for objects with different heights. For example, Figure~\ref{intro} (c) illustrates this problem. The representative  method RadarNet falls short in the detection performance for tall objects -- the truck class even experiences 0.5\% AP degradation after fusing the Radar data. 
In order to better harvest the benefit of LiDAR and Radar fusion, our viewpoint is that Radar features need to be more powerful before being fused. Therefore, we first enrich the Radar features -- with the help of LiDAR data -- and then integrate the enriched Radar features into the LiDAR processing branch for more effective fusion. 
As depicted in Figure~\ref{intro} (b), we refer to this scheme as \emph{bi-directional} fusion. And in this work,  we introduce a framework, {\emph \Name}, to achieve this goal. 
Specifically, \Name first encodes BEV features for each modality individually.
Next, it engages the query-based LiDAR-to-Radar (L2R) height feature fusion and query-based L2R BEV feature fusion, in which we query and group LiDAR points and LiDAR BEV features that are close to the location of each non-empty gird cell on the Radar feature map, respectively. 
The grouped LiDAR raw points are aggregated to formulate pseudo-Radar height features, and the grouped LiDAR BEV features are aggregated to produce pseudo-Radar BEV features. The generated pseudo-Radar height and BEV features are fused to the Radar BEV features through concatenation.
After enriching the Radar features, \Name then performs the Radar-to-LiDAR (R2L) fusion in a unified BEV representation.
Finally, a BEV detection network consisting of a BEV backbone network and a detection head is applied to output 3D object detection results.
We validate the merits of bi-directional LiDAR-Radar fusion via evaluating our \Name on nuScenes and Oxford Radar RobotCar (ORR)~\cite{barnes2020oxford} datasets.
On nuScenes dataset, \Name improves the mAP($\uparrow$) by \textbf{2.7\%} and reduces the mAVE($\downarrow$) by \textbf{5.3\%} against the LiDAR-centric baseline CenterPoint~\cite{yin2021center}, and remarkably outperforms the strongest counterpart, \emph{i.e.}, RadarNet, in terms of AP by absolutely \textbf{2.0}\% for cars and \textbf{6.3}\% for motorcycles. Moreover, \Name generalizes well on the ORR dataset, which has a different Radar data format and achieves 1.3\% AP improvements for vehicle detection.

In summary, we make the following contributions:
\vspace{-3pt}
\begin{itemize}
   \vspace{-3pt}\item We propose a bi-directional fusion framework, namely Bi-LRFusion, to combine LiDAR and Radar features for improving 3D dynamic object detection. 

   \vspace{-3pt}\item We devise the query-based L2R height feature fusion and query-based L2R BEV feature fusion to enrich Radar features with the help of LiDAR data.

   
   
   %
   
  \vspace{-3pt}\item We conduct extensive experiments to validate the merits of our method and show considerably improved results on two different datasets.

  
\end{itemize}

\section{Related Work}


\subsection{3D Object Detection with Only LiDAR Data}
In the literature, 3D object detectors built on LiDAR point clouds are categorized into two directions, \emph{i.e.}, point-based methods and voxel-based methods. 

\begin{figure*}[!t]
\centering
\includegraphics[width=.99\textwidth]{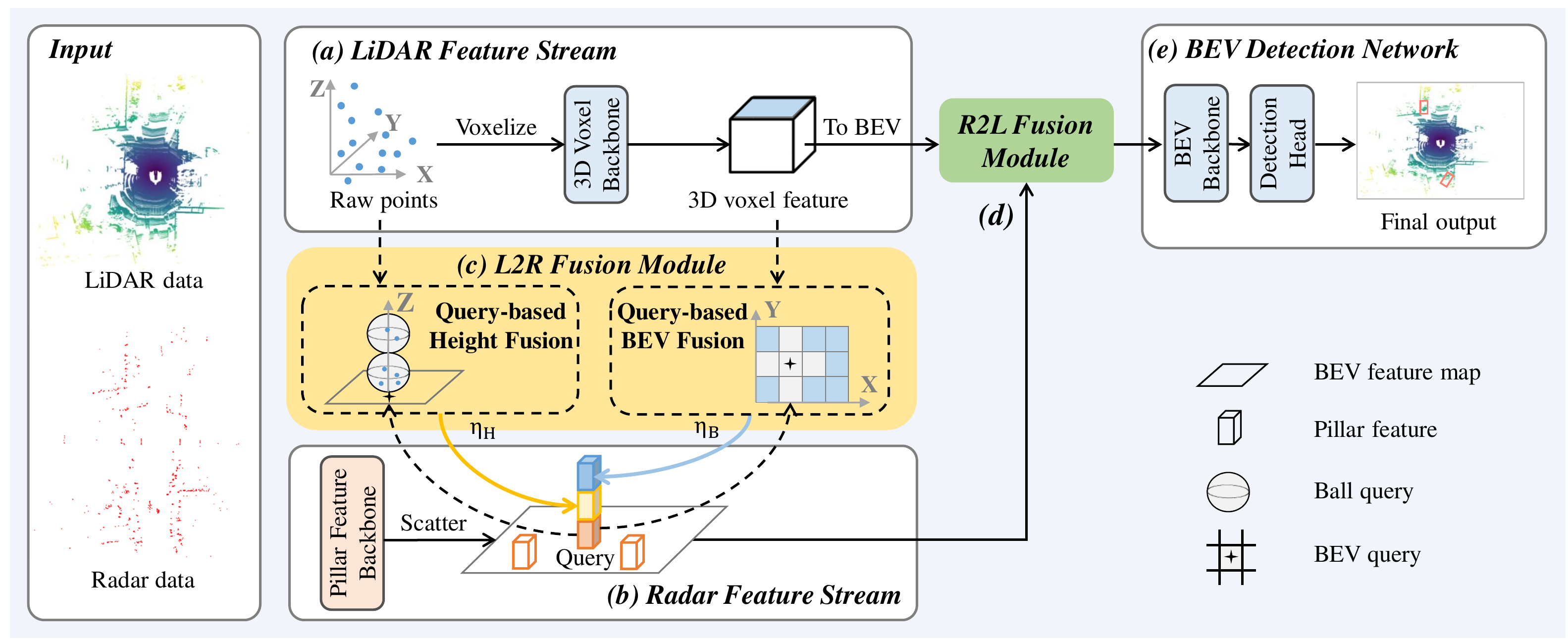}
\caption{An overview of the proposed \Name{ }framework. \Name includes five main components: (a) a LiDAR feature stream to encode LiDAR BEV features from LiDAR data, (b) a Radar feature stream to encode Radar BEV features from Radar data, (c) a LiDAR-to-Radar (L2R) fusion module composed of a query-based height feature fusion block and a query-based BEV feature fusion block, in which we enhance Radar features from LiDAR raw points and LiDAR features,  (d) a Radar-to-LiDAR (R2L) fusion module to fuse back the enhanced Radar features to the LiDAR-centric detection network, and (e) a BEV detection network that uses the features from the R2L fusion module to predict 3D bounding boxes for dynamic objects.
}

\vspace{-1.0em}
\label{network}
\end{figure*}

\vspace{4pt}\noindent{\textbf{Point-based Methods.}} Point-based methods maintain the precise position measurement of raw points, extracting point features by point networks\cite{qi2017pointnet,qi2017pointnet++} or graph networks~\cite{wang2019dynamic,wang2021object}.
The early work PointRCNN~\cite{shi2019pointrcnn} follows the two-stage pipeline to first produce 3D proposals from pre-assigned anchor boxes over sampled key points, and then refines these coarse proposals with region-wise features.
 STD~\cite{yang2019std} proposes the sparse-to-dense strategy for better proposal refinement.
A more recent work 3DSSD~\cite{yang20203dssd} further introduces feature based key point sampling strategy as a complement of previous distance based one, and develops a one-stage object detector operating on raw points.

\vspace{4pt}\noindent{\textbf{Voxel-based Methods.}} Voxel-based methods first divide points into regular grids, and then leverage the convolutional neural networks~\cite{zhou2018voxelnet,yan2018second,lang2019pointpillars,deng2021voxel,deng2021multi,shi2021pv} and Transformers~\cite{mao2021voxel,fan2022embracing} for feature extraction and bounding box prediction. 
SECOND\cite{yan2018second} reduces the computational overhead of dense 3D CNNs by applying sparse convolution.
PointPillars~\cite{lang2019pointpillars} introduces a pillar representation (a particular form of the voxel) to formulate with 2D convolutions.
To simplify and improve previous 3D detection pipelines, CenterPoint~\cite{yin2021center} designs an anchor-free one-stage detector, which extracts BEV features from voxelized point clouds to find object centers and regress to 3D bounding boxes.
Furthermore, CenterPoint introduces a velocity head to predict the object's velocity between consecutive frames.
In this work, we exploit CenterPoint as our LiDAR-only baseline. 

\subsection{3D Object Detection with Radar Fusion}
Providing larger detection range and additional velocity hints, Radar data show great potential in 3D object detection. However, since they are too sparse to be solely used~\cite{caesar2020nuscenes}, Radar data are generally explored as the complement of RGB images or LiDAR point clouds. The approaches that fuse Radar data for improving 3D object detectors can be summarized into two categories: one is input-level fusion, and the other is feature-level fusion.

\vspace{4pt}\noindent\textbf{Input-level Radar Fusion.}
RVF-Net~\cite{nobis2021radar} develops an early fusion scheme to treat raw Radar points as additional input besides LiDAR point clouds, ignoring the differences in data property between Radar and LiDAR.
As studied in ~\cite{cheng2021robust}, these input-level fusion methods directly incorporate Radar raw information into LiDAR branch, which is sensitive to even slight changes of the input data and is also unable to fully utilize the muli-modal feature.

\vspace{4pt}\noindent\textbf{Feature-level Radar Fusion.}
The early work GRIF~\cite{kim2020grif} proposes to extract region-wise features from Radar and camera branches, and to combine them together for robust 3D detection. Recent works generally transform image features to the BEV plane for feature fusion~\cite{huang2021bevdet,li2022bevformer,harley2022simple}.
MVDNet~\cite{qian2021robust} encodes  Radar points' intensity into a BEV feature map, and fuses it with LiDAR features to facilitate vehicle detection under foggy weather.
The representative work RadarNet~\cite{yang2020radarnet} first extracts LiDAR and Radar features via modality-specific branches, and then fuses them on the shared BEV perspective. 
The existing feature-level Radar fusion methods commonly ignore the problems caused by the height missing and extreme sparsity of Radar data, as well as overlooking the information intensity gap when fusing the multi-modal features.

Despite both input-level and feature-level methods have
improved 3D dynamic object detection via Radar fusion, they follow the uni-directional fusion scheme.
On the contrary, 
our \Name, for the first time, treats LiDAR-Radar fusion in a bi-directional way. We enhance the Radar feature  with the help of LiDAR data to alleviate issues caused by the absence of height information and extreme sparsity, and then fuse it to the LiDAR-centric network to achieve further performance boosting.





\section{Methodology}
In this work, we present~\Name, a bi-directional LiDAR-Radar fusion framework for 3D dynamic object detection. As illustrated in Figure~\ref{network}, the input LiDAR and Radar points are fed into the sibling LiDAR feature stream and Radar feature stream to produce their BEV features. 
Next, we involve a LiDAR-to-Radar (L2R) fusion step to enhance the extremely sparse Radar features that lack discriminative details. 
Specifically, for each valid (\emph{i.e.}, non-empty) grid cell on the Radar feature map, we query and group the nearby LiDAR data (including both raw points and BEV features) to obtain more detailed Radar features. Here, we focus on the height information that are completely missing in the Radar data and the local BEV features that are scarce in the Radar data. Through two query-based feature fusion blocks, we can transfer the knowledge from LiDAR data to Radar features, leading to much-enriched Radar features.  
Subsequently, we perform Radar-to-LiDAR (R2L) fusion by integrating the enriched Radar features to the LiDAR features in a unified BEV representation.
Finally, a BEV detection network composed of a BEV backbone network and a detection head outputs 3D detection results. Below we describe these steps one by one.


\subsection{Modality-Specific Feature Encoding}
\label{sec:feature_enc}

\noindent \textbf{LiDAR Feature Encoding.}
LiDAR feature encoding consists of the following steps.
Firstly, we divide LiDAR points into 3D regular grids and encode the feature of each grid by a multi-layer perception (MLP) followed by max-pooling. Here, a grid is known as a voxel in the discrete 3D space, and the encoded feature is the voxel feature. After obtaining voxel features, we follow the common practice to exploit a 3D voxel backbone network composed of 3D sparse convolutional layers and 3D sub-manifold convolutional layers~\cite{yan2018second} for LiDAR feature extraction.
Then, the output feature volume is stacked along the $Z$-axis, producing a LiDAR BEV feature map $M_{L}\in \mathbb{R} ^{ C_{1}\times H\times W}$, where $(H,W)$ indicate the height and width.

\vspace{4pt}\noindent \textbf{Radar Feature Encoding.}
A Radar point contains the 2D coordinate ($x$, $y$), the Radar cross-section $rcs$, and the timestamp $t$. 
In addition, we may also have\footnote{Due to the different 2D millimeter wave Radar sensors installed, the Radar data in the ORR dataset are stored in the form of intensity images and therefore do not include the other features mentioned here.}the radial velocity of the object in $X-Y$ directions ($v_{X},v_{Y}$), the dynamic property $dynProp$, the cluster validity state $invalid$\_$state$, and the false alarm probability $pdh0$. 
Please note that the Radar point's value on the $Z$-axis is set to be the Radar sensor's height by default. Here, we exploit the pillar~\cite{lang2019pointpillars} representation to encode Radar features, which directly converts the Radar input to a pseudo image in the bird's eye view (BEV).
Then we extract Radar features with a pillar feature network, obtaining a Radar BEV feature map $M_{R}\in \mathbb{R} ^{ C_{2}\times H\times W}$. 

\subsection{LiDAR-to-Radar Fusion}
\label{sec:l2r}
LiDAR-to-Radar (L2R) fusion is the core step in our proposed Bi-LRFusion. It involves two L2R feature fusion blocks, in which we generate pseudo height features, as well as the pseudo local BEV features, by suitably querying the LiDAR features. These pseudo features are then fused into the Radar features to enhance their quality.

\vspace{4pt}\noindent \textbf{Query-based L2R Height Feature Fusion.} As illustrated in Figure~\ref{intro} (c), directly fusing the Radar features to LiDAR-based 3D detection networks may lead to unsatisfactory results, especially for objects that are taller than the Radar sensor. This is caused by the fact that Radar points do not contain the height measurements.
To address this insufficiency, we design the query-based L2R height feature fusion (QHF) block to transfer the height distributions from LiDAR raw points to Radar features.   


\begin{figure}[!t]
\centering
\includegraphics[width=0.47\textwidth]{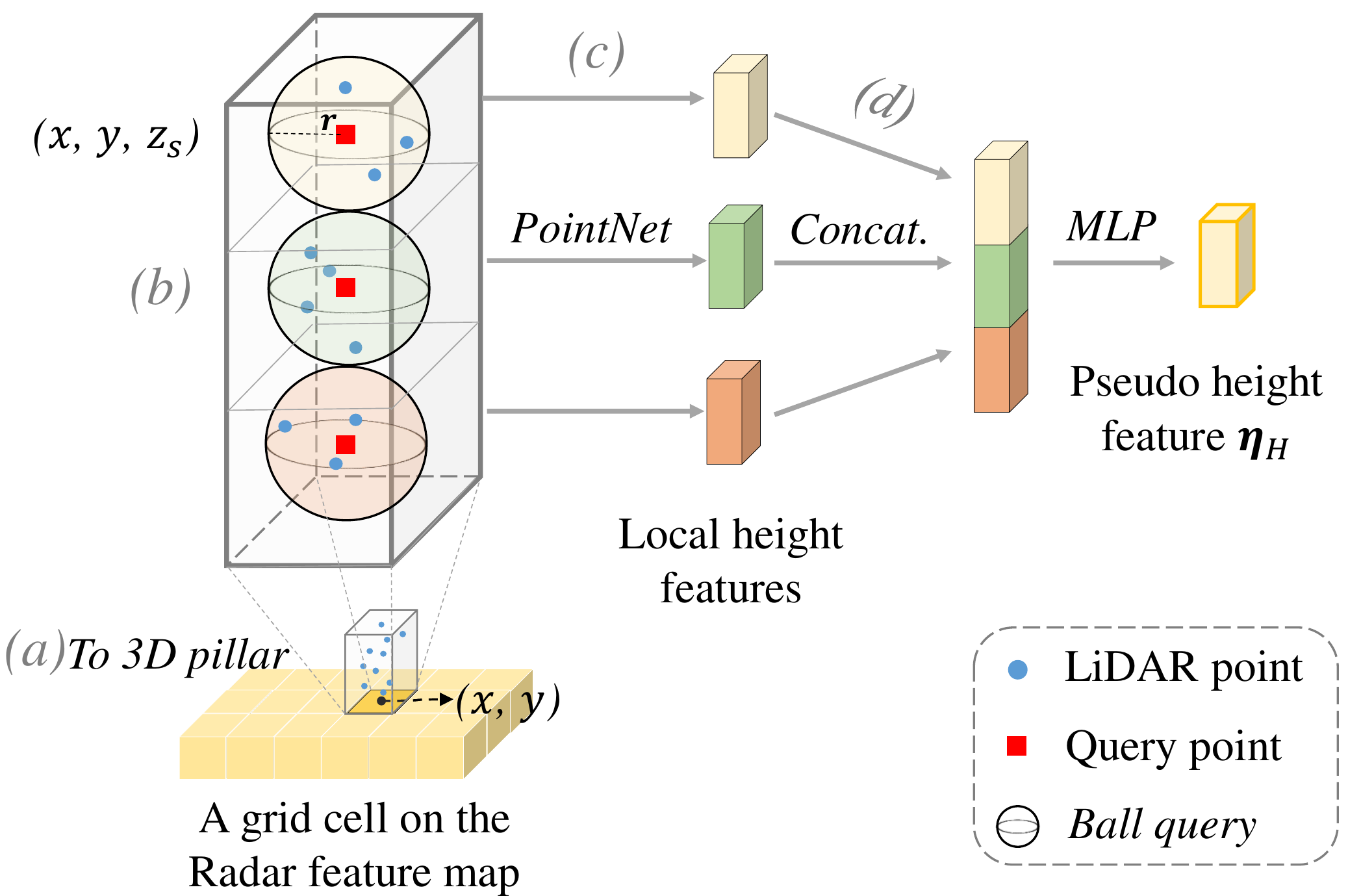}
\caption{An illustration of query-based L2R height feature fusion (QHF) block. QHF involves the following steps:
(a) lifting the non-empty grid cell on the Radar feature map to a pillar, and equally dividing the pillar into segments of different heights,
(b) querying and grouping neighboring LiDAR points based on the location of each segment's center,
(c) aggregating grouped LiDAR points to get the local height feature of each segment,
and (d) merging the segments' feature together to produce pseudo height feature $\boldsymbol{\eta}_{H}$ for the corresponding Radar grid cell.}
\vspace{-1.5em}
\label{lidar-guided}
\end{figure}

The pipeline of QHF is depicted in Figure \ref{lidar-guided}. The core innovation of QHF is the height feature querying mechanism. 
Given the $l$-th valid grid cell on the Radar feature map centered at ($x$, $y$), we first ``lift'' it from the BEV plane to the 3D space, which results in a pillar of height $h$.
Then, we evenly divide the pillar into $M$ segments along the height and assign a query point at each segment's center.
Specifically, let us denote the grid size of the Radar BEV feature map as $r \times r$. In order to query LiDAR points without overlap, the radius of the ball query is set to $r/2$, and the number of segments $M$ is set to $h/2r$. For the query point in $s$-th segment shown in Figure \ref{lidar-guided}, 
the ($x$, $y$) coordinates are from the central point of the given grid cell, which can be calculated with the grid indices, and grid sizes, together with the boundaries of the Radar points.
Further, the height value $z_{s}$ of the query point is calculated as:
\begin{align}
\mathrm {z}_{s}= z_{M} + r\times (2s - 1), 
\end{align}
where $z_{M}$ is the minimum height value among all the LiDAR points.
After establishing the query points, we then apply ball query~\cite{qi2017pointnet++} followed by a PointNet module~\cite{qi2017pointnet} to aggregate the local height feature $\mathbf {F }_{s}$ from the grouped LiDAR points. 
The calculation can be formulated as:
\begin{align}
\mathbf {F}_{s}= \max_{k = 1,2,\cdots,K}\left \{ \Psi ( {n}_{s}^{k}  )   \right \}, 
\end{align}
where ${n}_s^k$ denotes the $k$-th grouped LiDAR point in the $s$-th ball query segment, $K$ is the number of grouped points from ball query, $\Psi \left ( \cdot  \right ) $ indicates an MLP, and $\max \left ( \cdot  \right ) $ is the max pooling operation.
After obtaining the local height feature of each segment, we concatenate them together and feed the concatenated feature into an MLP to make the output channels have the same dimensions as the Radar BEV feature map. 
Finally, the output pseudo height feature $\boldsymbol{\eta}_{H}^{l}$ for the $l-$th grid cell on the Radar feature map produced by QHF is computed as:
\begin{align}
\boldsymbol{\eta}_{H}^{l}= \mathrm{MLP} \left ( \mathrm{Concat} \left ( \left \{ \mathbf {F }_{s} \right \}_{s=0}^{M}  \right )  \right ) .
\end{align}

\vspace{4pt}\noindent \textbf{Query-based L2R BEV Feature Fusion.}
When extremely sparse Radar BEV pixels are convoluted by convolutional kernels, the resulting Radar BEV feature barely retains valid local information since most of the neighboring pixels are empty.
To alleviate this problem, we design a query-based L2R BEV feature fusion block (QBF) that can generate the pseudo BEV feature which is more detailed than the original Radar BEV feature, by querying and grouping the corresponding fine-grained LiDAR BEV features. 
\begin{figure}[!t]
\centering
\includegraphics[width=0.48\textwidth]{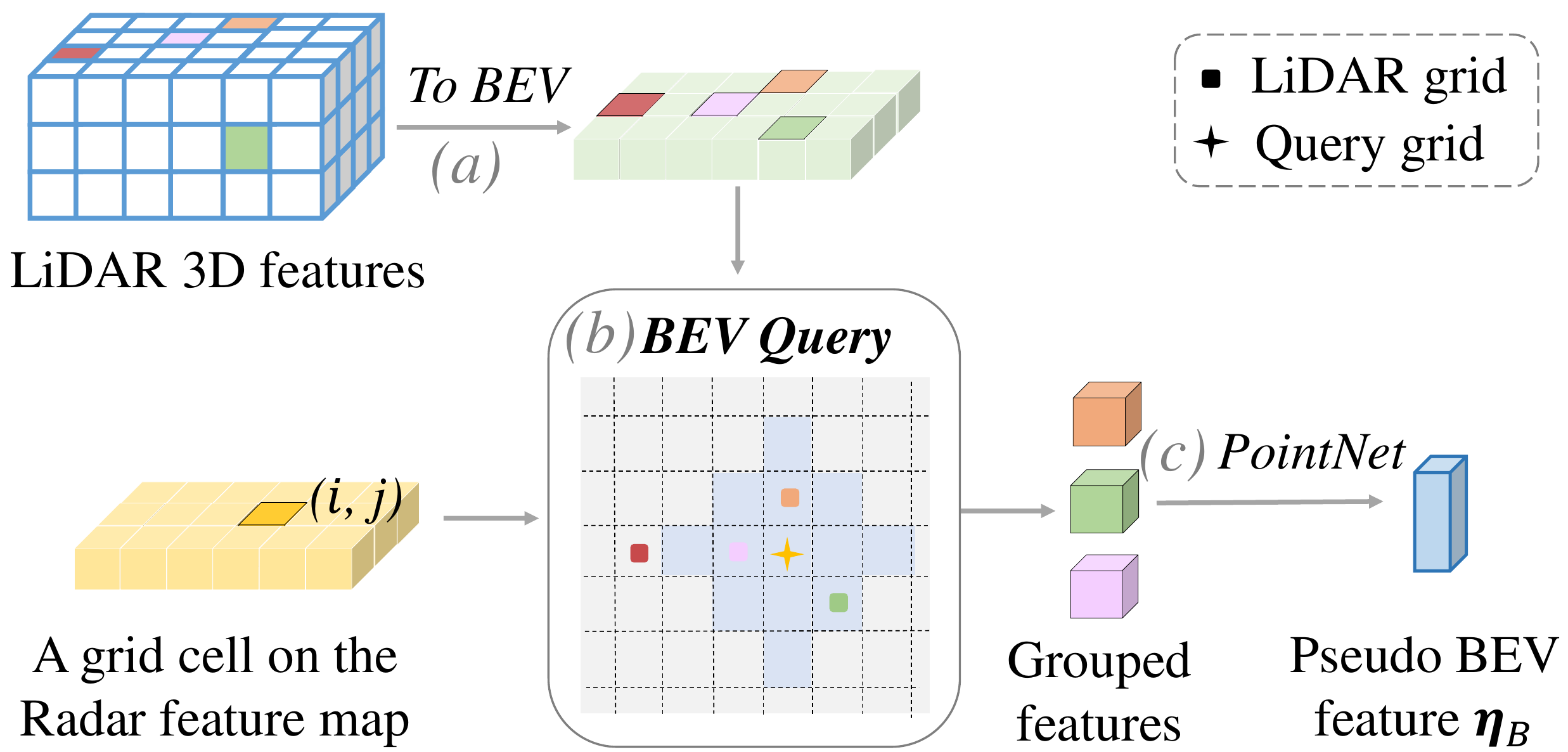}
\caption{
An illustration of query-based L2R BEV feature fusion (QBF) block with the following steps: 
(a) collapsing LiDAR 3D features to the BEV grids,
(b) querying and grouping LiDAR grids that are close to the Radar query grid based on their indices on the BEV,
and (c) aggregating the features of the grouped LiDAR grids to generate the pseudo BEV feature $\boldsymbol{\eta}_{B}$ for the corresponding Radar grid cell.
}
\vspace{-1.5em}
\label{lidar-guided2}
\end{figure}

The pipeline of QBF is depicted in Figure~\ref{lidar-guided2}. The core innovation of QBF is the local BEV feature querying mechanism that we describe below.
We first collapse LiDAR 3D features to the BEV grid plane~\cite{lang2019pointpillars}, forming a set of non-empty LiDAR grids (the same grid size as the Radar grid cell in Figure~\ref{lidar-guided2}). 
Given the $l$-th non-empty grid cell on the Radar feature map with indices ($i$, $j$),
we query and group LiDAR grid features that are close to the Radar grid cell on the BEV plane. 
Specifically, towards this goal, we propose a local BEV feature query inspired by Voxel R-CNN~\cite{deng2021voxel}, which finds all LiDAR grid features that are within a certain distance from the querying grid based on their grid indices on the BEV plane.
Specifically, we exploit the Manhattan distance metric and sample up to $K$ non-empty LiDAR grids within a specific distance threshold on the BEV plane. The Manhattan distance $D\left ( \alpha ,\beta  \right ) $ between indices of LiDAR grids $\alpha =\left \{ i_{\alpha },j_{\alpha  } \right \} $ and $\beta =\left \{ i_{\beta },j_{\beta  } \right \} $ can be calculated as:
\begin{align}
D\left ( \alpha ,\beta  \right ) & = \left | i_{\alpha }-i_{\beta } \right |+ \left | j_{\alpha }-j_{\beta } \right |,
\end{align}
where $i$ and $j$ are the indices of the LiDAR grid along the $X$ and $Y$ axis.
After the $l$-th BEV feature query, we obtain the corresponding features of the grouped LiDAR grids.
Finally, we apply a PointNet module~\cite{qi2017pointnet} to aggregate the pseudo Radar BEV feature $\boldsymbol{\eta}_{B}^{l}$, which can be summarized as:
\begin{align}
 \boldsymbol{\eta}_{B}^{l}= \max_{k = 1,2,\cdots,K}\left \{ \Psi^{'}  ({F}_{l}^{k}  ) \right \},
\end{align}
where ${F}_{l}^{k}$ denotes the $k$-th grouped LiDAR grid feature in the $l$-th BEV query mechanism. 

\subsection{Radar-to-LiDAR Fusion}
\label{sec:r2l}
After enriching the Radar BEV features with the pseudo height feature $\boldsymbol{\eta}_{H}$ and pseudo BEV features $\boldsymbol{\eta}_{B}$, we obtain the enhanced Radar BEV features that have 96 channels.

In this step, we fuse the enhanced Radar BEV features to the LiDAR-based 3D detection pipeline so as to incorporate valuable clues such as velocity information. Specifically, we concatenate the two BEV features in the channel-wise fashion following the practice from~\cite{yang2020radarnet,liu2022bevfusion}. 
Before forwarding to the BEV detection network, we also apply a convolution-based BEV encoder to help curb the effect of misalignment between multi-modal BEV features. The BEV encoder adjusts the fused BEV feature to 512 through three 2D convolution blocks.
\subsection{BEV Detection Network}
\label{sec:head}
Finally, the combined LiDAR and Radar BEV features are fed to the BEV detection network to output the results. The BEV detection network consists of a BEV network and a detection head.
The BEV network is composed of several 2D convolution blocks, which generate center features forwarding to the detection head.
As such, we use a class-specific center heatmap head to predict the center location of all dynamic objects and a few regression heads to estimate the object size, rotation, and velocity based on the center features.
We combine all heatmap and regression losses in one common objective and jointly optimize them following the baseline CenterPoint~\cite{yin2021center}.

\begin{table*}[!ht]
	\caption{
        \small{Comparison with other methods on the nuScenes validation set.
	We add ``$\ast $" to indicate that it is a reproducing version based on CenterPoint\cite{yin2021center}.
	We group the dynamic targets to (1) similar-height objects and (2) tall objects according to the Radar sensor's height.
	}
	}\label{tab:val}
	\vspace{-17pt}
    \begin{center}
        \addtolength\tabcolsep{3.8pt}
        \renewcommand\arraystretch{0.9}
        \small
		\scalebox{0.95}{
		\begin{tabular}{l|ccc|cccc|ccc}
		\hline
		\toprule[1.2pt]
            \multirow{2}{*}{Method} &\multirow{2}{*}{Modality} & \multirow{2}{*}{mAVE~$\downarrow$}  & \multirow{2}{*}{mAP~$\uparrow$}   &\multicolumn{4}{c|}{AP~$\uparrow$ of Group 1} & \multicolumn{3}{c}{AP~$\uparrow$ of Group 2} \\
             &                           &      &                    & Car& Motor. & Bicycle & Ped. & Truck & Bus & Trailer   \\ \midrule[0.8pt]
            PointPillars~\cite{lang2019pointpillars}&\multirow{3}{*}{L}&34.2&47.1&80.7&26.7&5.3& 70.8&49.4&62.0&34.9\\
            SECOND~\cite{yan2018second}&&32.1&52.9&81.7&40.1&18.2&77.4&51.9&66.0 &38.2 \\
            CenterPoint~\cite{yin2021center}&&30.3& 59.3  & 84.6 & 54.7 & 34.9 & 83.9 & 54.4 & 66.7 & 36.7 \\\midrule[0.8pt]
            RVF-Net~\cite{nobis2021radar} &\multirow{4}{*}{L + R}&- & 54.9 & - & - & - & - & - & - & -\\
            RadarNet~\cite{yang2020radarnet} & &- & - & 84.5 & 52.9 & - & - & - & - & -\\
            RadarNet$^\ast $~\cite{yang2020radarnet} &&26.2 & 60.4 & 85.9 & 57.4 & 38.4& 84.1 & 53.9 & 66.7& 37.2   \\
            \Name   &&$\bm{25.0}$ &$\bm{62.0}$  &$\bm{86.5}$  & $\bm{59.2}$ & $\bm{42.0}$ & $\bm{84.4}$ & $\bm{55.2}$ & $\bm{67.9} $&$\bm{38.4}$   \\
            \bottomrule[1.2pt]
		\end{tabular}
		}
	\end{center}
	\vspace{-2.0em}
\end{table*}

\section{Experiments}
We evaluate \Name on both the nuScenes and the Oxford Radar RobotCar (ORR) datasets and conduct ablation studies to verify our proposed fusion modules. We further show the advantages of \Name on objects with different heights/velocities.


\subsection{Datasets and Evaluation Metrics}

\vspace{4pt}\noindent \textbf{NuScenes Dataset.}
NuScenes~\cite{caesar2020nuscenes} is a  large-scale dataset for
3D detection including camera images, LiDAR points, and Radar points. 
We mainly adopt two metrics in our evaluation. The first one is the average precision ($AP$)  with a match threshold of 2D center distance on the ground plane. The second one is $AVE$ which stands for absolute velocity error in $m/s$ -- its decrease represents more accurate velocity estimation.
We average the two metrics over all 7 dynamic classes (mAP, mAVE), following the official evaluation.
%

\vspace{4pt}\noindent \textbf{ORR Dataset.}
The ORR dataset, mainly for localization and mapping tasks, is a challenging dataset including camera images, LiDAR points, Radar scans, GPS and INS ground truth.
We split the first data record into 7,064 frames for training and 1,760 frames for validation.
As this dataset does not provide ground truth of object annotation, we therefore follow the MVDet~\cite{qian2021robust} to generate 3D boxes of vehicles.
For evaluation metrics, we use AP of oriented bounding boxes in BEV to validate the vehicle detection performance via the CoCo evaluation framework~\cite{lin2014microsoft}.

\subsection{Implementation Details}
\vspace{4pt}\noindent \textbf{LiDAR Input.}
As allowed in the nuScenes submission rule, we accumulate 10 LiDAR sweeps to form a denser point cloud.
we set detection range to $\left ( -54.0,54.0 \right ) m$ for the $X$, $Y$ axis, and $\left ( -5.0,3.0 \right ) m$ for $Z$ axis with a voxel size of
$\left ( 0.075,0.075,0.2 \right) m $.
For ORR dataset, we set the point cloud range to $(-69.12, 69.12) m$ for $X$ and $Y$ axis, $(-5.0, 2.0) m$ for $Z$ axis and voxel size to $(0.32, 0.32, 7.0) m$.

\vspace{4pt}\noindent \textbf{Radar Input in NuScenes.}
Radar data are collected from 5 long-range Radar sensors and stored as BIN files, which is the same format as LiDAR point cloud.
We stack points captured by five Radar sensors into full-view Radar point clouds.
We also accumulate 6 sweeps to form denser Radar points.
The detection range of Radar data is consistent with the LiDAR range.
The voxel size is $\left ( 0.6, 0.6, 8.0 \right )m$ for nuScenes and $(0.32, 0.32, 7.0) m$ for ORR dataset.
We transfer the 2D position and velocity of Radar points from the Radar coordinates to the LiDAR coordinates.

\vspace{4pt}\noindent \textbf{Radar Input in ORR.}
The Radar data are collected by a vehicle equipped with only one spinning Millimetre-Wave FMCW Radar sensor and are saved as PNG files. 
Therefore, we need to convert the Radar images to Radar point clouds.
First, we use cen2019~\cite{8793990} as a feature extractor to extract feature points from Radar images. Attributes of every extracted point include $x$,$y$,$z$,$rcs$,$t$, where $rcs$ is represented by gray-scale values on the Radar image, and $z$ is set to 0.
To decrease the noise points and ghost points due to multi-path effects, we apply a geometry-probabilistic filter~\cite{pfilter} and the number of points is reduced to around 1,000 per frame.
Second, since Radar’s considerable scanning delay could cause the lack of synchronization with the LiDAR, we compensate for the ego-motion via SLAM~\cite{wang2022maroam}. 
Please refer to supplementary materials for more details.

\begin{table*}[!ht]
	\caption{
        \small{Comparison with other methods on the nuScenes test set. We add ``$\ast $'' to indicate that we reproduce and submit the result based on CenterPoint\cite{yin2021center}. We group the dynamic targets to (1) similar-height objects and (2) tall objects according to the Radar sensor's height.
	}
	}\label{tab:test}
    \begin{center}
        \addtolength
        \tabcolsep{3.8pt}
        \renewcommand\arraystretch{0.9}
        \small
        \vspace{-17pt}
		\scalebox{0.95}{
	\begin{tabular}{l|ccc|cccc|ccc}
		\hline
		\toprule[1.2pt]
            \multirow{2}{*}{Method} &\multirow{2}{*}{Modality} & \multirow{2}{*}{mAVE~$\downarrow$}  & \multirow{2}{*}{mAP~$\uparrow$}   &\multicolumn{4}{c|}{AP~$\uparrow$ of Group 1} & \multicolumn{3}{c}{AP~$\uparrow$ of Group 2} \\
             &                           &      &                    & Car& Motor. & Bicycle& Ped. & Truck & Bus & Trailer   \\ \midrule[0.8pt]
            PointPillars~\cite{lang2019pointpillars}&\multirow{6}{*}{L}&  31.6  & 30.5& 68.4 & 27.4 & 1.1& 59.7& 23.0 & 28.2 & 23.4   \\
            InfoFocus~\cite{wang2020infofocus} &&- &39.5 & 77.9 & 29.0 & 6.1& 64.3 & 31.4 & 44.8 & 37.3 \\
            PointPillars+~\cite{vora2020pointpainting}&& 27.0 & 40.1 & 76.0 & 34.2 & 14.0& 64.0 & 31.0 & 32.1 & 36.6 \\
            SSN~\cite{zhu2020ssn}&&26.6 & 42.6 & 82.4 & 48.9 & 24.6& 75.6 & 41.8 & 46.1 & 48.0  \\
            CVCNet~\cite{chen2020every}&& 26.8 & 55.3 & 82.7& 59.1 & 31.3 & 79.8 & 46.1 & 46.6 & 49.4 \\
            
            CenterPoint~\cite{yin2021center} &&28.8 & 60.3 & 85.2 & 59.5 & 30.7& 84.6 & 53.5 & 63.6 & 56.0  \\ \midrule[0.8pt]
            RadarNet$^\ast $~\cite{yang2020radarnet} &\multirow{2}{*}{L + R}& 27.9 & 61.9 & 86.0 & 59.6 & 31.9 & $\bm{84.8}$& 53.0 & 62.8 & 55.4 \\
            \Name   && $\bm{25.7}$& $\bm{63.1}$ & $\bm{87.2}$& $\bm{59.9}$  & $\textbf{34.0}$& 84.7 & $\bm{54.8}$ & $\bm{63.7}$ & $\bm{57.3}$     \\
            \bottomrule[1.2pt]
		\end{tabular}
		}
	\end{center}
	\vspace{-2.0em}
\end{table*}

\begin{table}[!ht]
	\caption{
        \small{Comparison with other methods of vehicle detection accuracy on the ORR dataset.
	}
	}\label{tab:ORR}
	\vspace{-17pt}
    \begin{center}
        \addtolength
        \tabcolsep{5.0pt}
        \renewcommand\arraystretch{0.9}
        \small
		\scalebox{0.95}{
		\begin{tabular}{l|c|cc}
		\hline
		\toprule[1.2pt]
            \multirow{2}{*}{Method} &\multirow{2}{*}{Modality} & \multicolumn{2}{c}{AP~$\uparrow$} \\ 
                         &                       & IoU=0.5 &  IoU=0.8  \\\midrule[0.8pt]
            PIXOR~\cite{yang2018pixor}&\multirow{2}{*}{L}&  72.8 & 41.2 \\
            PointPillars~\cite{lang2019pointpillars}&&85.7 & 58.3\\ \midrule[0.8pt]
            DEF~\cite{bijelic2020seeing} &\multirow{3}{*}{L + R}& 86.6 & 46.2 \\
            MVDet~\cite{qian2021robust} &&  90.9 &$\bm{74.6} $ \\
            \Name   &&  $\bm{92.2}$ & 74.4  \\
            \bottomrule[1.2pt]
		\end{tabular}
		}
	\end{center}
	\vspace{-2.0em}
\end{table}

\subsection{Comparison on NuScenes Dataset}
We evaluate our \Name with AP (\%) and AVE (m/s) of 7 dynamic classes on both validation and test sets.

First, we compare our \Name~with a few top-ranked methods on the nuScenes validation set, including the vanilla CenterPoint~\cite{yin2021center} as our baseline, the LiDAR-Radar fusion models and other LiDAR-only models on nuScenes benchmark.
For a fair comparison, we use consistent settings with the original papers and reproduce the results on our own.
Table~\ref{tab:val} summarizes the results.  
Overall, our results show solid performance improvements. 
The SOTA LiDAR-Radar fusion method, RadarNet~\cite{yang2020radarnet}, mainly focused on improving the AP values for cars and motorcycles in their study. Compared to RadarNet, we can further improve the AP by \textbf{+2.0\%} for cars and \textbf{+6.3}\% for motorcycles. 
In addition, when we consider all dynamic object categories, \Name~can increase the mAP($\uparrow$) by \textbf{+2.7\%} and improve mAVE($\downarrow$) by \textbf{-5.3\%} compared to CenterPoint. Meanwhile, our approach surpasses the reproduced RadarNet$^\ast$ by +1.6\% in mAP and -1.2\% in mAVE.

We also compare our method with several top-performing models on the test set.
The detailed results are listed in Table~\ref{tab:test}. The results show that \Name gives the best average results, in terms of both mAP and mAVE. For six out of  seven individual object categories, it has the best AP results. Even for the only exception category (the pedestrian), its AP is the second best, with only 0.1\% lower than the best. 
Overall, the results on val and test sets consistently demonstrate the effectiveness of \Name.

\begin{figure}[t]
	\centering
	\includegraphics[width=0.39\textwidth]{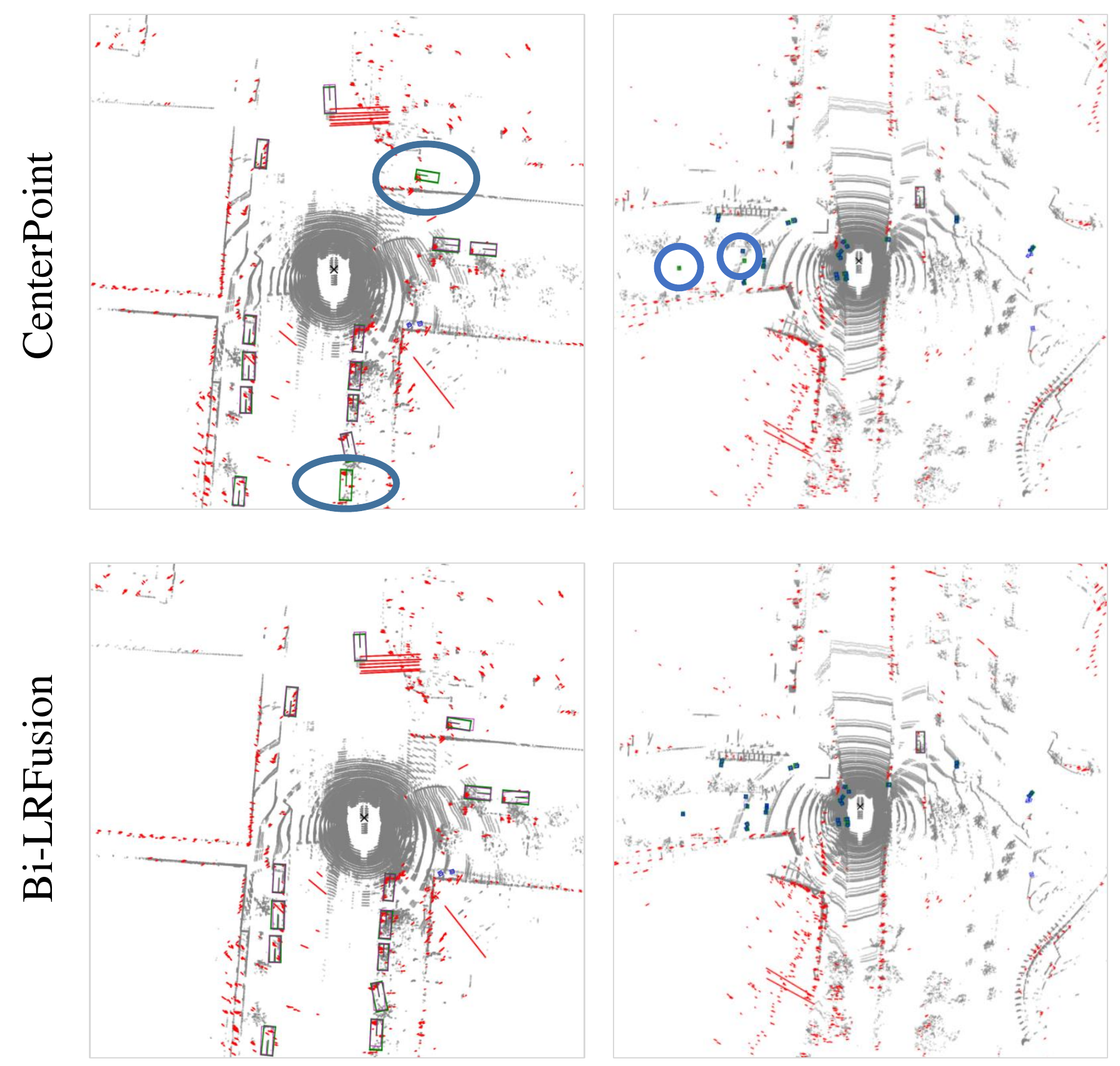}
	\caption{Qualitative comparison between the LiDAR-only method CenterPoint~\cite{yin2021center} and our \Name. 
	The grey dots and red lines are LiDAR points and Radar points with velocities.
	We visualize the prediction and ground truth (green) boxes.
	Specifically, blue circles represent the missed detections from CenterPoint but \Name corrects them by integrating Radar data.
	}

	\label{fig:vis}
	\vspace{-1.4em}
\end{figure}

Furthermore, we qualitatively compare with CenterPoint and \Name on the nuScenes dataset. Figure~\ref{fig:vis} shows a few visualized detection results, which demonstrates that the enhanced Radar data from L2R feature fusion module can indeed better eliminate the missed detections. 
\subsection{Comparison on ORR Dataset}
To further validate the effectiveness of our proposed \Name, we conduct experiments on the challenging ORR dataset. Due to the lack of  ground truth annotations, we only report the AP for cars with IoU thresholds of 0.5 and 0.8, following the common COCO protocol.

We compare our \Name with several LiDAR-only and LiDAR-Radar fusion methods on the ORR dataset.
From Table~\ref{tab:ORR},
\Name achieves the best AP results with IoU threshold of 0.5, and second best with IoU threshold of 0.8. For IoU threshold of 0.8, its AP is only 0.2\% lower than the best, and much higher than other schemes.   
Table~\ref{tab:ORR} also shows that \Name is effective in handling different Radar data formats.


\begin{table}[t!]
\centering
\caption{The effect of each proposed component in \Name. It shows that all components contribute to the overall detection performance. 
}
\vspace{-9pt}
\small
\addtolength\tabcolsep{0.01pt}
\renewcommand\arraystretch{0.9}
\resizebox{0.48\textwidth}{!}{%
\begin{tabular}{c|c|cc|cc}
\hline
\toprule[1.2pt]
\multirow{2}{*}{Method} & \multirow{2}{*}{R2L fusion}  & \multicolumn{2}{c|}{L2R fusion}                        & \multirow{2}{*}{mAP~$\uparrow$} & \multirow{2}{*}{mAVE~$\downarrow$ } \\ 
             &         & QHF               & \multicolumn{1}{c|}{QBF} &                      &                       \\ \midrule[0.8pt]
(a) & &  &  &59.3 &30.3  \\
(b) & \checkmark&  &    &{60.4}~(+1.1)  &{26.2}~(-4.1) \\ 
(c) &\checkmark&\checkmark  &   &{61.3}~(+2.0) & {25.1}~(-5.2)   \\
(d) &\checkmark&  & \checkmark   &{61.9}~(+2.6)  &{25.5}~(-4.8)   \\
(e) &\checkmark & \checkmark   & \checkmark   &$\bm{62.0}$~(\red{+2.7})  &$\bm{25.0}$~(\red{-5.3})   \\
 \bottomrule[1.2pt]
\end{tabular}
}
\label{tab:ablation_table2}
\vspace{-1.5em}
\end{table}

\subsection{Effect of Different \Name Modules}
To understand how each module in \Name affects the detection performance, we conduct ablation studies on the nuScenes validation set and report its mAP and mAVE of all dynamic classes in Table \ref{tab:ablation_table2}.

\emph{Method (a)} is our LiDAR-only baseline CenterPoint, which achieves mAP of 59.3\% and mAVE of 30.3\%. 

\emph{Method (b)} extends \emph{(a)} by simply fusing the Radar feature via R2L fusion, which improves mAP by 1.1\% and mAVE by -4.1\%. This indicates that integrating the Radar feature is effective to improve 3D dynamic detection.

\emph{Method (c)} extends \emph{(b)} by utilizing the raw LiDAR points to enhance the Radar features via query-based L2R height feature fusion (QHF), which leads to an improvement of 2.0\% mAP and -5.2\% mAVE. 

\emph{Method (d)} extends \emph{(b)} by taking advantage of the detailed LiDAR features on the BEV plane to enhance Radar features via query-based L2R BEV feature fusion (QBF), improving mAP by 2.6\% and mAVE by -4.8\%. 

\emph{Method (e)} is our \Name. By combining all the components, it achieves a gain of 2.7\% for mAP and -5.3\% for mAVE compared to CenterPoint. By enhancing the Radars features before fusing them into the detection network, our bi-directional LiDAR-Radar fusion framework can considerably improve the detection of moving objects.
\begin{table}[t]
\caption{mAP and mAVE results for LiDAR-only CenterPoint~\cite{yin2021center} and \Name~for objects with different velocities.}
\vspace{-7pt}
    \centering
    \begin{subtable}[t]{\linewidth}
\small
\addtolength\tabcolsep{1.5pt}
\renewcommand\arraystretch{0.9}
\resizebox{0.98\textwidth}{!}{
\begin{tabular}{l| c | c  c  }
\hline
\toprule[1.2pt]
Method & Velocity (m/s) & mAP~$\uparrow$ & mAVE~$\downarrow$ \\
\midrule[0.8pt]
\multirow{4}{*}{\begin{tabular}[c]{@{}l@{}}
CenterPoint \end{tabular}} & $0 - 0.5$                      & 81.3          & 7.0                       \\
& $0.5 -5.0 $                      & 44.9          & 57.1                       \\
& $5.0 - 10.0$                      & 61.5          & 62.7                       \\
& $\geq10.0$                      & 55.4          & 83.5                       \\  \midrule[0.8pt]
\multirow{4}{*}{\begin{tabular}[c]{@{}l@{}}\Name \end{tabular}}& 
$0 - 0.5$                      & \ress{82.6}{+1.3}          & \res{6.9}{-0.1}                       \\

& $0.5 - 5.0$                       & \ress{52.1}{+7.2}           & \res{45.9}{-11.2}                         \\

& $5.0 - 10.0$                      & \ress{68.6}{+7.1}             & \res{51.4}{-11.3}                         \\

& $\geq 10.0$                      & \ress{65.6}{+10.2}             & \res{69.7}{-13.8}                             \\
 \bottomrule[1.2pt]
\end{tabular}}
\caption{cars}
\end{subtable}

\begin{subtable}[t]{\linewidth}
\small
\addtolength\tabcolsep{1.5pt}
\renewcommand\arraystretch{0.9}
\centering
\resizebox{0.98\textwidth}{!}{
\begin{tabular}{l| c | c c }
\hline
\toprule[1.2pt]
Method & Velocity (m/s) & mAP~$\uparrow$ & mAVE~$\downarrow$ \\
\midrule[0.8pt]
\multirow{4}{*}{\begin{tabular}[c]{@{}l@{}}CenterPoint\end{tabular}} & $0-0.5$                      & 52.7          & 6.2                       \\
& $0.5-5.0$                      & 31.0          & 82.4                       \\
& $5.0-10.0$                      & 33.8          & 115.1                       \\
& $\geq 10.0$                      & 54.0          & 255.5                       \\ \midrule[0.8pt]

\multirow{4}{*}{\begin{tabular}[c]{@{}l@{}}\Name \end{tabular}} & 
$0-0.5$                      & \ress{54.0}{+1.3}          & \res{6.0}{-0.2}                       \\
& $0.5-5.0$                      & \ress{37.3}{+6.3}           & \res{79.1}{-3.3}                         \\
& $5.0-10.0$                      & \ress{42.5}{+8.7}             & \res{93.4}{-21.7}                         \\
& $\geq$10.0                      & \ress{70.2}{+16.2}             & \res{154.8}{-100.7}                             \\
 \bottomrule[1.2pt]
\end{tabular}}
        \caption{motorcycles}
    \end{subtable}
    \label{tab:velo}
    \vspace{-1.0em}
\end{table}

\begin{table}[t]

\centering
\caption[STH]{The percentage (\%) of objects with different velocities in nuScenes dataset, \emph{i.e.,} stationary (0 m/s - 0.5 m/s), low-velocity (0.5 m/s - 5.0 m/s), medium-velocity (5.0 m/s - 10.0 m/s), high-velocity ($\ge$ 10.0 m/s). These groups are divided according to \cite{caesar2020nuscenes}.}
\vspace{-7pt}
\small
\addtolength\tabcolsep{1.0pt}
\renewcommand\arraystretch{0.6}
\resizebox{0.45\textwidth}{!}{
\begin{tabular}{l|cccc}
\hline
\toprule[1.2pt]
Class                       & $0 - 0.5$ & $0.5 - 5.0$ & $5.0 - 10.0$ & $\geq 10.0$ \\
                                 \midrule[0.8pt]
Cars                    & 72.6                & 11.7                  & 11.8                   & 3.9                              \\
Motorcycles             & 69.8                & 13.0                  & 12.3                   & 4.9                              \\ 
 \bottomrule[1.2pt]
\end{tabular}
}
\label{tab:ratio}
\vspace{-1.4em}
\end{table}



\subsection{Effect of Object Parameters}
We also evaluate the performance gain of \Name for objects with different heights/velocities compared with the LiDAR-centric baseline CenterPoint.

\vspace{3pt}\noindent\textbf{Effect of Object Height.}
We group the dynamic objects 
into two groups: (1) regular-height objects including cars, motorcycles, bicycles, pedestrians, and (2) tall objects including trucks, buses, and trailers.
Note that since millimeter wave is not sensitive to non-rigid targets, the AP gain of pedestrians is hence small or even nonexistent.
From Table~\ref{tab:val}, RadarNet$^\ast $ shows +1.9\% AP for group (1) and +0.0\% AP for group (2) averagely over the CenterPoint, while \Name shows \textbf{+3.5\%} AP for group (1) and \textbf{+1.2\%} AP for group (2) averagely.
We summarize that RadarNet$^\ast $ improves the AP values for group (1) objects, but not for group (2).
The height missing problem in each Radar feature makes it difficult to detect specific tall objects.
On the other hand, our \Name effectively avoids this problem and improves the AP for both group (1) and (2) over the baseline CenterPoint, regardless of the object's height.


\vspace{3pt}\noindent\textbf{Effect of Object Velocity.}
Next, we look at the advantage of \Name~over objects that move at different speeds. 
Following~\cite{caesar2020nuscenes}, we consider objects with speed$>$0.5m/s as moving objects, 5m/s and 10m/s are the borderlines to distinguish low and medium moving state. 
More concretely, we also show the percentages of cars and motorcycles within different velocity ranges in the nuScenes dataset in Table~\ref{tab:ratio}: around 70\% of cars and motorcycles are stationary, only around 4\% of them are moving faster than $>$10m/s, and the rest of them are uniformly distributed in the low / medium velocity range. 
Tables~\ref{tab:velo} (a) and (b) show the mAP and mAVE of the LiDAR-only detector and \Name~ for cars and motorcycles that move at different speeds. 
The results show that the performance improvement is more pronounced for objects with high velocity. 
This demonstrates the effectiveness of Radar in detecting dynamic objects, especially those with faster speeds. 


\section{Conclusion}

In this paper, we introduce \Name, a bi-directional fusion framework that fuses the complementary LiDAR and Radar data for improving 3D dynamic object detection.
Unlike existing LiDAR-Radar fusion schemes that directly integrate Radar features into the LiDAR-centric pipeline, we first make Radar features more discriminative by transferring the knowledge from LiDAR data (including both raw points and BEV features) to the Radar features, which alleviates the problems with Radar data, \emph{i.e.}, lack of object height measurements and extreme sparsity.
With our bi-directional fusion framework, \Name outperforms the earlier schemes for detecting dynamic objects on both nuScenes and ORR datasets.

\noindent\textbf{Acknowledgements.} We thank DequanWang for his help. This work was done during her internship at Shanghai AI Laboratory. This work was supported by the Chinese Academy of Sciences Frontier Science Key Research Project ZDBS-LY-JSC001, the Australian Medical Research Future Fund MRFAI000085, the National Key R\&D Program of China(No.2022ZD0160100), and Shanghai Committee of Science and Technology(No.21DZ1100100). 

{\small
\bibliographystyle{ieee_fullname}
\bibliography{reference}
}

\end{document}


\title{Bi-LRFusion: Bi-Directional LiDAR-Radar Fusion for 3D  Dynamic \\Object Detection\\
Supplementary Material}  

\author{Yingjie Wang$^{1}$\hspace{0.15cm} Jiajun Deng$^{2\ast}$\hspace{0.15cm} Yao Li$^{1}$\hspace{0.15cm} Jinshui Hu$^{3}$\hspace{0.15cm} Cong Liu$^{3}$\hspace{0.15cm} Yu Zhang$^{1}$\hspace{0.15cm}  Jianmin Ji$^{1}$ \\Wanli Ouyang$^{4}$\hspace{0.15cm} Yanyong Zhang$^{1\ast}$\hspace{0.15cm} \\
\normalsize $^1$University of Science and Technology of China \hspace{0.5cm} $^2$University of Sydney \hspace{0.5cm} $^3$iFLYTEK \hspace{0.5cm} $^4$Shanghai AI Laboratory\\
}

\maketitle
\thispagestyle{empty}
\appendix

\section{Overview}
The supplementary document is organized as follows:
\begin{itemize}

\item{Section~\ref{detail} presents the detailed model settings and training details.}

\item{Section~\ref{vis} presents additional qualitative results of two typical locations of Radar points on objects with different heights.} 

\item{Section~\ref{QHF} depicts the detailed effect of query-based height feature fusion (QHF) block (Section 4.4 of the main paper).} 

\item{Section~\ref{QBF} depicts the detailed effect of query-based BEV feature fusion (QBF) block (Section 4.4 of the main paper).} 

\item{Section~\ref{velo} depicts the effect on objects at different ranges.} 

\item{Section~\ref{sota1} depicts the additional comparison with existing methods.}

\end{itemize}
\section{Model Settings and Training Details}
\label{detail}
Our implementation is based on the open-sourced code mmdetection3d~\cite{mmdet3d2020}. 
We choose CenterPoint~\cite{yin2021center} and PointPillars~\cite{lang2019pointpillars} to serve as the baseline for the LiDAR-only detection on nuScenes and ORR datasets, respectively.
We follow the network of PointPillars for the Radar feature stream.
The radius of each ball query is 0.15$m$.
The Manhattan distance threshold is set as $\left (2,2\right) $ for BEV query.
The final output channels of the PointNet module and the MLP layer are 32.

Our implementation is based on the open-sourced code mmdetection3d. 
We choose CenterPoint and PointPillars to serve as the baseline for the LiDAR-only detection on nuScenes and ORR datasets, respectively.
We follow the network of PointPillars for the Radar feature stream.
The radius of each ball query is 0.15$m$.
The Manhattan distance threshold is set as $\left (2,2\right) $ for BEV query.
The final output channels of the PointNet module and the MLP layer are 32.

\section{Qualitative Results of Radar Point Locations on Objects with Different Heights}
\label{vis}
As mentioned in the main paper, the currently available Radar is unable to capture the height information. The height value of a Radar point is assigned as the height of the ego Radar sensor, which is deployed on the top of the autonomous vehicle. As a result, Radar points are located at different part for objects with different heights.
We illustrate the location of Radar points on object with different heights in Figure~\ref{vis2}. This treatment leads to two typical types of Radar point locations: (1) For cars and motorcycles whose actual heights are similar to the data-collecting car, Radar points often fall on the top of their bounding boxes, and (2) for much taller objects such as trucks and buses, Radar points instead fall inside of their bounding boxes. As discussed in the Figure 1 from the main paper, the missing height information of Radar leads to unstable improvements for objects with different height.

\section{Effect of QHF Block}
\label{QHF}
We further report the mean AP 
of two dynamic object groups on nuScenes validation set to illustrate how QHF block affects the detection performance.
As proposed in the main paper, we group the dynamic objects to (1) similar-height objects and (2) tall objects according to the Radar sensor's height.
From Table~\ref{tab:QHF}, \emph{Method (a)} is our LiDAR-only baseline CenterPoint~\cite{yin2021center}.
\emph{Method (b)} extends \emph{(a)} by simply fusing the Radar feature via R2L fusion, which achieves 1.9\% gain of AP for group (1). We notice that tall objects in group (2) do not enjoy as much performance gain. 
As such, \emph{Method (c)} enriches Radar data by learning the corresponding height features from raw LiDAR points via QHF block, leading to steady AP improvement for both two groups (2.3\%/1.3\%). 
It demonstrates that pseudo height information from QHF plays a more important role in improving the detection performance of large objects. 
\begin{figure}[t]
	\centering
	\includegraphics[width=0.48\textwidth]{fig/vis2.pdf}
	\caption{
	Due to the lack of height information, Radar points may fall at different locations on an object's bounding box depending upon the actual height of the object.
	The middle part of the figure is an 2D image from nuScenes dataset~\cite{caesar2020nuscenes}; in dotted circles, we show the 3D ground-truth bounding boxes and Radar points using MeshLab~\cite{cignoni2008meshlab} tool. 
    Best viewed in color.
	}

	\label{vis2}
	\vspace{-1.4em}
\end{figure}

\section{Effect of QBF Block}
We also present the detailed performance gain of QBF block in Table~\ref{tab:QHF}.
\emph{Method (d)} extends \emph{(b)} by learning from detailed LiDAR features on the BEV plane to enhance Radar features, which achieves +2.3\% improvements in terms of AP for Group 1 and 0.4\% for Group 2.
The designed QBF block constructs a more fine-grained Radar BEV feature for more effective detection, benefiting from much-detailed clues, \emph{e.g.} objects at long range or objects with small sizes.  
In addition, we can observe that \emph{Method (d)} only brings small performance gains on tall objects.
We speculate that this is mainly due to the height missing problem still exists when using the QBF block alone.
\label{QBF}

\section{Effect of Objects at Different Ranges}
\label{velo}
Apart from object parameters like heights and velocities, we also evaluate the improvements of \Name for objects with different distances compared with the LiDAR-centric baseline CenterPoint.
Table~\ref{range} shows the mAP and mAVE of the LiDAR-only detector and \Name on the representative car class at different ranges.
At the range closer than 20 m, CenterPoint performs better due to the precise and dense point cloud from LiDAR data.
At the range farther than 20 m, the performance of our \Name exceeds the LiDAR-only method with the advantage of a long detection range from Radar data.
The result demonstrates that our LiDAR-Radar fusion framework is less sensitive to the distance because the LiDAR points become sparse as the distance increases, while the Radar points are uniform compared to the LiDAR points.
In conclusion, Radar data are of great help to locating and detecting targets especially at long distances.

\begin{table}[t]
	\caption{
        {
     The detailed effect of each proposed component in \Name.
    We report the mean AP on two typical groups. Group (1) includes similar-height objects like cars, motorcycles, bicycles and pedestrians, while group (2) includes tall objects like trucks, buses and trailers.
    We also show the performance gain of different components compared with LiDAR-only method.
	}
	\vspace{-0.6em}
	}\label{tab:QHF}
        \addtolength\tabcolsep{-1pt}
        \renewcommand\arraystretch{1.2}
        \small
		\scalebox{0.82}{
		\begin{tabular}{c|c|cc|cc|ccc}
		\hline
		\toprule[1.2pt]
            \multirow{2}{*}{Method}  & \multirow{2}{*}{R2L }& \multicolumn{2}{c|}{L2R } &\multicolumn{2}{c|}{Mean AP~$\uparrow$ (\%)} & \multicolumn{2}{c}{Gain (\%)} \\
             &  & QHF&    \multicolumn{1}{c|}{QBF}         & Group 1& Group 2 & Group 1 & Group 2  \\ \midrule[0.8pt]
            
            (a)&&&&64.5& 52.6  & - & -  \\ 
            (b) &\checkmark&&&{66.4}& {52.6} &\red{+1.9} & \red{+0.0} \\

            (c) &\checkmark&\checkmark &&{66.8}& {53.9} & \red{+2.3} & \red{+1.3}  \\
         
            (d) &\checkmark&&\checkmark&{68.6}& {53.0} & \red{+4.1} & \red{+0.4}  \\

            \bottomrule[1.2pt]
		\end{tabular}
		}
\end{table}

\begin{table}[t]
\caption{
mAP and mAVE results for LiDAR-only CenterPoint~\cite{yin2021center} and \Name~for cars at different ranges.
}\vspace{-0.6em}
    
\footnotesize
\renewcommand\arraystretch{1.2}
\resizebox{0.48\textwidth}{!}{
        \begin{tabular}{l|l|ccc}
		\toprule[1.2pt]
\multirow{2}{*}{Method}                                                             & Range& \multicolumn{3}{c}{\emph{AP}(\%) on Car}                                        \\
                                                                                    &      (m)                      & 0.5m        & 1.0m          & 2.0m                  \\ \midrule[0.8pt]
\multirow{3}{*}{\begin{tabular}[c]{@{}l@{}}LiDAR-only\end{tabular}} & 0-20                      & 93.3          & 96.5          & 96.5                 \\
                                                                                    & 20-40                     & 69.8          & 82.2          & 85.5                  \\
                                                                                    & $>$40                   & 26.1          & 40.1          & 45.8                  \\ \midrule[0.8pt]
\multirow{3}{*}{\begin{tabular}[c]{@{}l@{}}\Name \end{tabular}}       & 0-20                      & 92.6          & 95.5          & \res{96.6}{+0.1}          \\
                                                                                    & 20-40                     & \res{71.0}{+1.2}   & \res{83.5}{+1.3} & \res{86.7}{+1.2}  \\
                                                                                    & $>$40                     & \res{27.9}{+1.8}  & \res{42.6}{+2.5}  & \res{48.0}{+2.2}   \\
                                                                                    \bottomrule[1.2pt]
\end{tabular}}
\label{range}
\end{table}

\section{Effect of Sensor Misalignment levels}
\label{misalign}
We first define misalignment levels followed BEVFormer~\cite{li2022bevformer}.
For the $i$-th level, a translation noise sampled from a normal distribution $\sigma\sim(0, 2.5i)$ is introduced to each direction. Besides, a horizontal rotation noise sampled from a normal distribution $\sigma\sim(0, i)$ is also introduced.
From Table~\ref{alignment}, \Name is able to tolerate misalignment levels 1-3, thanks to the query-based mechanism that can effectively query features within a certain area (instead of a single point or a small number of points).
However, when the misalignment level increases beyond the query range, the performance starts to drop significantly (level 4). 
We will elaborate on the misalignment tolerance issue in the revision. 



\begin{table}[h]
\centering
\caption{ The mAP (\%) under different misalignment levels. }
\vspace{-0.6em}
\label{alignment}
    \renewcommand\arraystretch{1.2}
        \small
    \resizebox{0.48\textwidth}{!}{
    \addtolength\tabcolsep{8.5pt}
    \begin{tabular}{*{10}{l|c|c|c|c|c}}
        \toprule[1.2pt]
        Level & 0  & 1 & 2 & 3 & 4\\
        \midrule
        mAP (\%) &  62.0 & 61.8 & 61.5 & 61.4 & 60.8\\
        \bottomrule[1.2pt]
    \end{tabular}
    }
    \vspace{-1em}
\end{table}

\section{Comparison with existing methods}
\label{sota1}
Bi-LRFusion can be applied to different LiDAR-only detectors, not limited to CenterPoint.
Therefore, we choose TransFusion-L~\cite{bai2022transfusion}, which is also a prevalent baseline in the community, to conduct further experiments. As shown in Table~\ref{sota}, our Bi-LRFusion improves TransFusion-L considerably, demonstrating its generalizability.


\begin{table}[h]
\vspace{-1em}
\centering
\caption{Experimental Results with Transfusion-L. }
\label{sota}
    \vspace{-0.6em}
    \renewcommand\arraystretch{1.2}
        \small
    \resizebox{0.48\textwidth}{!}{
    \addtolength\tabcolsep{8.5pt}
    \begin{tabular}{*{10}{l|c|c|c}}
        \toprule[1.2pt]
       Methods & Modality & mAVE $\downarrow$ &  mAP $\uparrow$ \\
        
        \midrule
       
         Transfusion-L & L & 24.8 & 65.3  \\
        \quad + \Name & L+R & \textbf{23.7}& \textbf{67.5} \\
        \bottomrule[1.2pt]
    \end{tabular}
    }
    \vspace{-1em}
\end{table}

{\small
\bibliographystyle{ieee_fullname}
\bibliography{reference}
}